\documentclass[conference]{IEEEtran}
\usepackage{cite}
\usepackage{amsmath,amssymb,amsfonts}
\usepackage{graphicx}
\usepackage{textcomp}
\usepackage{xcolor}
\usepackage{mathptmx}
\usepackage[caption=false,font=footnotesize]{subfig}
\usepackage{bm}
\usepackage{algorithm}
\usepackage{algpseudocode} 
\usepackage{multirow}

\def\BibTeX{{\rm B\kern-.05em{\sc i\kern-.025em b}\kern-.08em
    T\kern-.1667em\lower.7ex\hbox{E}\kern-.125emX}}
\begin{document}

\newif\ifreview
\newif\iflong

\title{TimeREISE: Time-series Randomized Evolving Input Sample Explanation}

\ifreview
\author{\IEEEauthorblockN{Anonymous}}
\else
\author{
\IEEEauthorblockN{1\textsuperscript{st} Dominique Mercier}
\IEEEauthorblockA{\textit{Smart Data \& Knowledge Services} \\
\textit{DFKI GmbH}\\
Kaiserslautern, Germany \\
dominique.mercier@dfki.de}
\and
\IEEEauthorblockN{2\textsuperscript{nd} Andreas Dengel}
\IEEEauthorblockA{\textit{Smart Data \& Knowledge Services} \\
\textit{DFKI GmbH}\\
Kaiserslautern, Germany \\
andreas.dengel@dfki.de}
\and
\IEEEauthorblockN{3\textsuperscript{rd} Sheraz Ahmed}
\IEEEauthorblockA{\textit{Smart Data \& Knowledge Services} \\
\textit{DFKI GmbH}\\
Kaiserslautern, Germany \\
sheraz.ahmed@dfki.de}
}
\fi

\maketitle

\begin{abstract}
Deep neural networks are one of the most successful classifiers across different domains. However, due to their limitations concerning interpretability their use is limited in safety critical context. The research field of explainable artificial intelligence addresses this problem. However, most of the interpretability methods are aligned to the image modality by design. The paper introduces TimeREISE a model agnostic attribution method specifically aligned to success in the context of time series classification. The method shows superior performance compared to existing approaches concerning different well-established measurements. TimeREISE is applicable to any time series classification network, its runtime does not scale in a linear manner concerning the input shape and it does not rely on prior data knowledge.
\end{abstract}


\section{Introduction}
\label{sec:introduction}
The success of deep neural network is funded by the superior performance and scaling deep neural networks offer compared to traditional machine learning methods~\cite{allam2019big}. However, during the last decades the need of explainable decisions has become more important. In critical infrastructures it is inconceivable to use approaches without any justification of the results~\cite{peres2020industrial}. In the medical sector, financial domain, and other safety-critical areas explainable computations are required by law~\cite{bibal2020impact}. Furthermore, there are ethical constraints that limit the use of artificial intelligence even more~\cite{karliuk2018ethical, perc2019social}. Consequently, a large research domain evolved. This domain covers the explainable artificial intelligence (XAI). One major goal is to propose techniques that provide interpretable results to enable the broader use of deep neural networks.

For several years researcher developed modifications of the networks and model agnostic methods to provide these results~\cite{dovsilovic2018explainable}. The majority of these methods originates from the image modality as its concepts are easier to interpret for humans~\cite{zhang2018visual}. Especially, model agnostic methods have shown great success. One famous category of model agnostic approaches are attribution methods~\cite{das2020opportunities}. The number of available methods of this category increases every year. One advantage of them is their loose coupling with the network. In addition, they do not limit the processing capabilities of the network. Although, some attribution methods come up with small limitations concerning the network architecture. The downside of these methods is that the provided results require additional human inspection and interpretation. Furthermore, they do not make any statement related to the concepts covered by the network. Revealing the concepts learned by the network is not the goal of these approaches. Considering the time series modality this is not a huge drawback as concepts are not well defined in this domain and an explanation based on pre-defined concepts would be not suitable. 

Despite their great success and the concept independence, not all of these methods can be applied to time series. Besides the above-mentioned limitations additional properties arise in the time series context. These properties are less important for the image modality but they are pivotal for the success of an attribution method in the time series context. Noisy explanations are acceptable in the image domain but can results in low information gain when it comes to time series interpretability. Another aspect is the Continuity of the attribution~\cite{abdul2020cogam}. It is pivotal for time series attributions that a certain degree of Continuity is preserved in the explanation. An explanation that suffers for large spikes of important data points within small windows introduces ambiguity and cognitive load. Due to the possible infinite length and number of channels it is unavoidable to focus on every data point. The explanation needs to highlight the important time frames and channels. This is not the case in the image domain as the number of channels and their role is predefined. The channels in the image domain are used together which is not possible in the time series domain.

Taking into account above-mentioned limitations and time series specific properties there is no perfect attribution methods available for time series. This paper proposes TimeREISE, an instance-based attribution method applicable to every classifier. It addresses common bottlenecks such as runtime, smoothness, and robustness against input perturbations as mentioned in~\cite{mercier2022time}. The rest of the paper shows that the explanations provided by TimeREISE are continuous, precise and robust. Without prior knowledge about the dataset it is possible to produce attribution methods with different granularity and smoothness. The approach is inspired by RISE~\cite{petsiuk2018rise} and different perturbation-based attribution methods. Two major advantages are the following: TimeREISE can be applied to backbox classifiers and its runtime does not scale directly with the input shape of the data.

\section{Related Work}
\label{sec:related}
Interpretability methods are wide spread across the different modalities such as image, natural language, and time series. A good overview of the diversity of these methods is given by Das and Rad~\cite{das2020opportunities}. Independent of the modality the goal is to identify an important subset of features to overcome ethical and industrial restrictions as mentioned by Peres et al.~\cite{peres2020industrial} and Karliuk~\cite{karliuk2018ethical}. One prominent class of interpretability methods are attribution techniques. 

The first sub category of attribution methods covers the gradient-based approaches. A good survey of these was provided by Anacona et al.~\cite{ancona2019gradient}. These methods use the backpropagation to compute the importance of the features. Speaking of the advantages and disadvantages of these methods, they are known for their superp runtime but suffer from the noisy gradients and the access to the model internals. Guided-Backpropagation and IntegratedGradients are two well-known gradient-based methods. The Guided-backpropgation computes the gradient with respect to the target prediction based on the non-negative gradients. More information about this approach was provided by Sundararajan et al.~\cite{sundararajan2017axiomatic}. IntegratedGradients uses so called baselines and approximates the integral of the gradients compared to the baseline. Further information is given by Springerberg et al.~\cite{springenberg2014striving}. 

In contrast to these methods the perturbation-based techniques do not require full access to the model as they perturb the input. An disadvantages of these methods is the increase in time as they utilize multiple forward passes. One famous example is the FeatureAblation presented in Fisher et al.~\cite{fisher2019all}. Therefore, the features are replaced with a baseline value such as the mean and the prediction is used to evaluate the impact. Very similar to this approach is the Occlusion presented in Zeiler et al.~\cite{zeiler2014visualizing}. The features are removed completely. 

The last category covers methods that do not fit directly to the previously mentioned. One method that falls into this category is LIME, introduced by Ribeiro et al.~\cite{lime}. Although LIME performs perturbations to the input it is different in a way that a local model is trained to estimate the importance. 

To evaluate the effectiveness of an attribution maps a set of well-known metrics evolved. An important fact is that the ground truth of the feature importance is not given in most cases and the measurements have to deal with that. One approach is to perform a deletion and insertion test are two well-known techniques to evaluate the efficiency of an attribution methods. E.g. Petsiuk et al.~\cite{petsiuk2018rise} used them to provide evidence for their attribution method. Another well-known approach is to use the Infidelity and Sensitivity proposed by Yeh et al.~\cite{yeh2019fidelity}. To compute the Infidelity the attribution is perturbed by a significant amount and the change in the prediction is evaluated. In contrast to that, the Sensitivity perturbs the input by an insignificant amount and the attribution is compared to the original one. A third metric related to the robustness is the Continuity. A continuous attribution map may suffer in the insertion evaluation, however, smooth attribution maps are more robust against attacks. Detailed information about the adversarial robustness was given by Alvarez et al. ~\cite{alvarez2018robustness}. In addition smooth attribution maps require less cognitive effort for interpretation as stated by Abdul et al.~\cite{abdul2020cogam}, however the correctness of the method needs to be preserved~\cite{adebayo2018sanity}. Finally, one of the most important aspects is the scaling with respect to the runtime as this defines the usability.

\section{TimeREISE}
\label{sec:timereise}
\begin{figure}[!t]
\centering
\includegraphics[width=\linewidth]{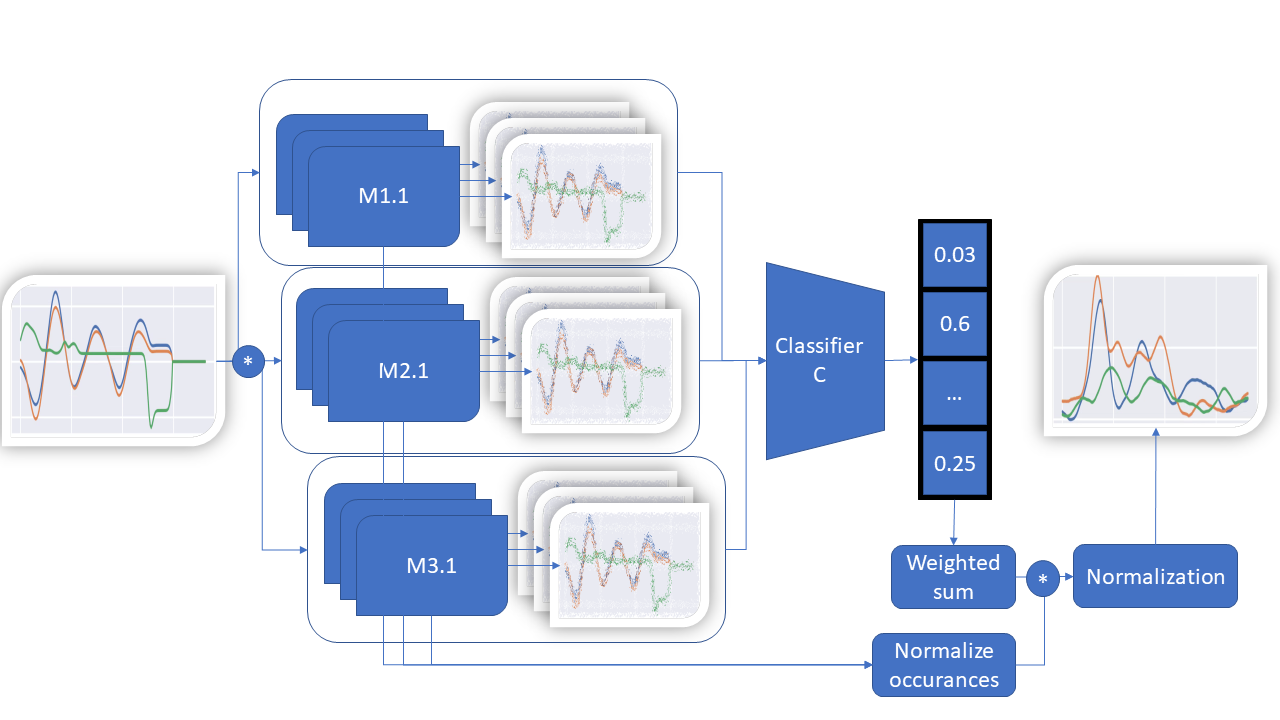}
\caption{\textbf{TimeREISE.} A set of masks with different density and granularity is applied to the input using an exchangeable perturbation function. The default perturbation is an elementwise multiplication. The masked input is passed to a classifier and the classification score is retrieved. The classification score is multiplied with the masks and normalized by the number of feature occurrences. Finally, the attribution is normalized to scores between zero and one.}
\label{fig:pipeline}
\end{figure}

This paper presents the novel approach TimeREISE a post-hoc interpretability methods applicable of any classification network. The work was inspired by Petsiuk et al.~\cite{petsiuk2018rise}. They presented a random perturbation based approach for the image domain that is used as baseline to build TimeREISE. Similar to RISE~\cite{petsiuk2018rise} masks are generated, applied to the input and the output confidence is measured using the classification scores. However, there are several adaptations in the native RISE~\cite{petsiuk2018rise} to enhance the approach and successfully apply it to time series data. Besides the simple normalization based on the occurrences of each data point TimeREISE was extended to create masks that evaluate the different channels. The second major addition applied is the summation over different probabilities. RISE~\cite{petsiuk2018rise} uses only a fixed probability of occluded points to create the masks resulting in a fixed density. In contrast to that, TimeREISE uses masks of different density and combines them in an additive manner which removes the assumption of the number of relevant data points. Figure~\ref{fig:pipeline} shows the overall workflow of TimeREISE. 

\subsection{Mathematical Formulation}
TimeRISE extends the native mathematical formulation presented by Petsiuk et al.~\cite{petsiuk2018rise} utilizing the different channels. TimeREISE generates masks with the shape $s'=(c, t')$ instead of $s''=(1, t')$ where $t'$ refers to the downsampled time axis and $c$ to the channels. This enhances TimeREISE to apply masks that occlude different timesteps $t'$ across all channels $c$ within a mask $s'$ instead of using the same timesteps $t'$ across all channels $c$ as it is the case for $s''$. Furthermore, the Monte Carlo sampling is performed across a set of densities $P$ and granularities $G$. This enhances the masks to consider serveral densitiy values $p$ to regularize the densitiy of the attribution. Similarly, the use of several granularity values $g$ regularizes the size of the occluded patches. This changes the set of masks as shown in Equation~\ref{eq:mask_set}. 

\begin{equation}
\label{eq:mask_set}
    M = \{ M^{p,g}_{0},\dots M^{p,g}_{N} \mid p \in P \wedge g \in G \}
\end{equation}

Finally, denote $S$ as the weighted sum of the scores produced by the network and the random masks $M$ similar to Petsiuk et al. but normalize each feature as shown in Equation~\ref{eq:feature_norm}.

\begin{equation}
    \label{eq:feature_norm}
    S = \sum_{c=0}^{C} \sum_{t=0}^{T} \frac{ S_{c,t} }{ \sum_{m=0}^{N} M_{c,t, n} }
\end{equation}

\subsection{Runtime Evaluation}
For the runtime evaluation the initialization and the attribution are considered as two separate processes. Equation~\ref{eq:runtime_mask_generation} shows the runtime to create the set of masks for a given set of density probabilities $P$, granularities $G$ and the number of masks $N$ defined for each combination of $p_{i}$ and $g_{i}$. $\beta$ is defined as the constant time to create the given map. In addition, $P$ and $G$ are independent of the data shape and therefore do not increase the can be considered as constant factors leading to a runtime of $\Theta (N)$.

\begin{equation}
    \label{eq:runtime_mask_generation}
    t_{init} = P * G * N * \beta \rightarrow t_{init} = \Theta (P * G * N) \rightarrow t_{init} = \Theta (N)
\end{equation}

Equation~\ref{eq:runtime_mask_application} shows the linear runtime of the attribution step. $\gamma$ is defined as the constant time to apply the perturbation and $\delta$ as the constant time the classifier requires to forward pass the sample. Similar to the initialization step $P$ and $G$ are assumed as constants which results in a runtime of $\Theta (N)$.

\begin{equation}
    \label{eq:runtime_mask_application}
    t_{apply} = P * G * N * \gamma * \delta \rightarrow t_{apply} = \Theta (N)
\end{equation}

\subsection{Theoretical Implementation}
The implementation of TimeREISE can be divided into two parts similar to the RISE~\cite{petsiuk2018rise} implementation by Petsiuk et al.~\cite{petsiuk2018rise}. In a first stage shown in Algorithm~\ref{alg:mask_generation} a set of masks suited for the input shape gets generated. This has to be executed only once per dataset. Therefore, consider every combination of probabilities $P$ and granularities $G$ provided. The probability refers to a threshold used to determine the density of the mask. The granularity refers to the amount of data that is considered in a single slice. The downsampling and upsampling are performed along the time axis. Uniform refers to a uniform distribution with the given shape $s'$. An additional cropping step is performed to preserve the original shape $s$.

Algorithm~\ref{alg:mask_application} performs the actual attribution. A predefined perturbation method $\sigma$ is applied to the input $x$ using every mask $m_{i}$ and is passed to the classifier $\theta$. As default perturbation the method uses the simple elementwise multiplication of the input $x$ and the mask $m_{i}$ as proposed by Petsiuk et al. This results in a list of scores stored in $S$. Next the matrix product of $S_{T}$ and the masks $M$ is computed and each point is normalized by the number of occurrences $N$ in the set $M$. Finally, the map is normalized to values between zero and one.

\begin{algorithm}[!t]
	\caption{Mask generation - Initialization}
	\label{alg:mask_generation}
	\begin{algorithmic}[1]
	    \State Define: $s$ as input shape, $P$ as set of probabilities, $G$ as set of granularities for time steps, $N$ number of masks, and $M$ as list of masks.
	    \For {$p=1,\ldots,P$}
			\For {$g=1,\ldots,G$}
			    \For {$i=1,\ldots,N$}
    			    \State $s' = downsample(s, g)$
    			    \State $m = uniform(s') < p$
    			    \State $m = upsamle(m, s)$
    			    \State $m = crop(m, s)$
    			    \State Append $m$ to $M$
    	        \EndFor
			\EndFor
        \EndFor
        \State $S = \frac{S \times M}{N}$
        \State $S = \frac{S - \min(S)}{\max(S - \min(S))}$
	\end{algorithmic} 
\end{algorithm}

\begin{algorithm}[!t]
	\caption{Mask application - Attribution} 
	\label{alg:mask_application}
	\begin{algorithmic}[1]
	    \State Define: $x$ as input, $\theta$ as classifier, $\sigma$ as perturbation function, $S$ as list of scores, and $N$ as feature occurrences across all masks $M$.
	    \For {$m=1,\ldots,M$}
    		\State $x_{m_{i}} = \sigma(x,m_{i})$
    		\State $y' = \theta(x_{m_{i}})$
    		\State Append $y'$ to $S$
        \EndFor
		\State $S = \frac{S^{T} \times M}{N}$
        \State $S = \frac{S - \min(S)}{\max(S - \min(S))}$
	\end{algorithmic} 
\end{algorithm}

\section{Datasets}
\label{sec:datasets}
The work uses multiple datasets from the well-known UEA \& UCR repository~\cite{tsc2021datasets} to perform the experiments. The selection of datasets is based on a sufficient number of samples and the dataset modalities such as number of time steps, channels, and classes. Furthermore, the list of datasets is extended using the Anomaly dataset proposed by Siddiqui et al.~\cite{siddiqui2019tsviz}. This synthetic dataset serves as an interpretable baseline as the point anomalies in this dataset are mathematically defined and therefore the ground truth attribution is available. Conversely, this is not the case for the other datasets and only limited interpretablility is given. Table~\ref{tab:datasets} lists the datasets and their characteristics. Supplementary, these are assigned to the critical infrastructure domains they belong to.

\begin{table}[!t]
\renewcommand{\arraystretch}{1.3}
\caption{\textbf{Datasets} related to critical infrastructures. Different characteristics such as the datasetsize, length, feature number and classes are covered by this selection.} 
\label{tab:datasets}
\centering
\begin{tabular}{l|r|r|r|r|r}
\textbf{Domain \& Dataset} & \textbf{Train} & \textbf{Test} & \textbf{Steps} & \textbf{Chls.} & \textbf{Cls.} \\
\hline\hline
\textbf{Critical Manufacturing} & & & & & \\
Anomaly                 & 35,000  & 15,000  & 50      & 3   & 2 \\
ElectricDevices         & 8,926   & 7,711   & 96      & 1   & 7 \\
FordA                   & 3,601   & 1,320   & 500     & 1   & 2 \\
\hline
\textbf{Food and Agriculture} & & & & & \\
Crop                    & 7,200   & 16,800  & 46      & 1   & 24 \\
Strawberry              & 613     & 370     & 235     & 1   & 2 \\
\hline
\textbf{Public Health} & & & & & \\
ECG5000                 & 500     & 4,500   & 140     & 1   & 5 \\
FaceDetection           & 5,890   & 3,524   & 62      & 144 & 2 \\
MedicalImages           & 381     & 760     & 99      & 1   & 10 \\
NonInvasiveFetalECG     & 1,800   & 1,965   & 750     & 1   & 42 \\
PhalangesOutlinesCorrect & 1800   & 858     & 80      & 1   & 2 \\
\hline
\textbf{Communications} & & & & & \\
CharacterTrajectories   & 1,422   & 1,436   & 182     & 3   & 20 \\
HandOutlines            & 1,000   & 370     & 2,709   & 1   & 2 \\
UWaveGestureLibraryAll  & 896     & 3,582   & 945     & 1   & 8 \\
Wafer                   & 1,000   & 6,164   & 152     & 1   & 2 \\
\hline
\textbf{Transportation Systems} & & & & & \\
AsphaltPavementType     & 1,055   & 1,056   & 1,543   & 1   & 3 \\
AsphaltRegularity       & 751     & 751     & 4,201   & 1   & 2 \\
MelbournePedestrian     & 1,194   & 2,439   & 24      & 1   & 10 \\
\end{tabular}
\end{table}

\section{Experiments}
\label{sec:experiments}
In the following paragraph describes the general setup to reproduce the results and cover decision that affect the experiments. Following to the generic experiment the paper provides experiments on the insertion and deletion of data based on the importance scores of the attribution methods, an Infidelity and sensitivity analysis and visual examples of the method and other state-of-the-art attribution methods.

As model InceptionTime the current state-of-the-art proposed by Pawaz et al.~\cite{fawaz2020inceptiontime} was used. The network is trained using a learning rate scheduler to half the learning rate on plateaus and early stopping to prevent overfitting. As optimizer SGD was used with an initial learning rate of $0.01$ and a maximum of $100$ epochs. As some datasets are very large and the computation of measures such as the Sensitivity is computationally very expensive, this work randomly sampled a set of $100$ test samples to perform the attribution on a representative subset. In addition, the base accuracy scores for the whole datasets and the subset are provided in Table~\ref{tab:baseline_accs}. Highlighting that the findings based on the subset can be transferred to the complete datasets. Concerning the attribution methods  GuidedBackprop~\cite{springenberg2014striving}, IntegratedGradients~\cite{sundararajan2017axiomatic}, FeatureAblation~\cite{zeiler2014visualizing}, Occlusion~\cite{zeiler2014visualizing}, and LIME~\cite{lime} were used as state-of-the-art methods. This set of methods cover all categories of attribution methods mentioned in Section~\ref{sec:related}.

\begin{table}[!t]
\renewcommand{\arraystretch}{1.3}
\caption{\textbf{Performance} of IncpetionTime. COncerning the accuracy, and f1 scores the subsampled dataset achieves similar performance and can be used as a set of representative samples for the further experiments.} 
\label{tab:baseline_accs}
\centering
\scalebox{0.82}{
\begin{tabular}{l|r|r|r|r|r|r}
\multirow{2}{*}{\textbf{Dataset}} & \multicolumn{3}{c|}{\textbf{Test data}} & \multicolumn{3}{c}{\textbf{100 samples}} \\
                         & macro  & micro  &  acc   & macro  & micro  &  acc   \\
\hline\hline
Anomaly                  & 0.9769 & 0.9871 & 0.9872 & 0.9699 & 0.9797 & 0.9800 \\
AsphaltPavementType      & 0.9169 & 0.9244 & 0.9242 & 0.8905 & 0.8991 & 0.9000 \\
AsphaltRegularity        & 1.0000 & 1.0000 & 1.0000 & 1.0000 & 1.0000 & 1.0000 \\
CharacterTrajectories    & 0.9940 & 0.9944 & 0.9944 & 1.0000 & 1.0000 & 1.0000 \\
Crop                     & 0.7189 & 0.7189 & 0.7281 & 0.7058 & 0.7228 & 0.7400 \\
ECG5000                  & 0.5611 & 0.9352 & 0.9436 & 0.6045 & 0.9412 & 0.9500 \\
ElectricDevices          & 0.6286 & 0.6935 & 0.7056 & 0.6709 & 0.7602 & 0.7900 \\
FaceDetection            & 0.6634 & 0.6634 & 0.6637 & 0.6779 & 0.6790 & 0.6800 \\
FordA                    & 0.9492 & 0.9492 & 0.9492 & 0.9294 & 0.9299 & 0.9300 \\
HandOutlines             & 0.9464 & 0.9510 & 0.9514 & 0.9399 & 0.9493 & 0.9500 \\
MedicalImages            & 0.7227 & 0.7461 & 0.7474 & 0.7086 & 0.7479 & 0.7500 \\
MelbournePedestrian      & 0.9422 & 0.9424 & 0.9422 & 0.9635 & 0.9595 & 0.9600 \\
NonInvasiveFetalECG      & 0.9400 & 0.9430 & 0.9425 & 0.8424 & 0.9240 & 0.9200 \\
PhalangesOutlinesCorrect & 0.8142 & 0.8254 & 0.8275 & 0.8849 & 0.8898 & 0.8900 \\
Strawberry               & 0.9554 & 0.9593 & 0.9595 & 0.9672 & 0.9699 & 0.9700 \\
UWaveGestureLibraryAll   & 0.9165 & 0.9167 & 0.9174 & 0.8525 & 0.8696 & 0.8700 \\
Wafer                    & 0.9954 & 0.9982 & 0.9982 & 1.0000 & 1.0000 & 1.0000 \\
\hline\hline
Average                  & 0.8613 & 0.8911 & 0.8931 & 0.8593 & 0.8954 & 0.8988 \\
\end{tabular}
}
\end{table}

\subsection{Sanity Check: Insertion \& Deletion}
The causal metric was used by Fong and Vedaldi~\cite{fong2017interpretable} to explain the importance values of an attribution method. The intuition behind the the deletion is that the prediction of a classifier changes if the cause of the actual class within the sample get removed. This applies to the insertion as well. In the case of the deletion the points starting with the most important one are removed from the input and the prediction is computed. Large drops suggest that the feature was significant for the prediction. Further, the AUC based on the sequential deletion of features to rank the methods across every dataset were computed. In case of the deletion lower AUCs suggest that the method is superior in spotting important parts of the input. Similar, the same was done for the insertion starting with a sample that has only mean values. For the insertion higher AUCs are superior. Large increases in this setup correspond to adding important data points relevant for the prediction. 

\begin{figure}[!t]
\centering
\subfloat[Deletion of important data points]{
\includegraphics[width=.96\linewidth]{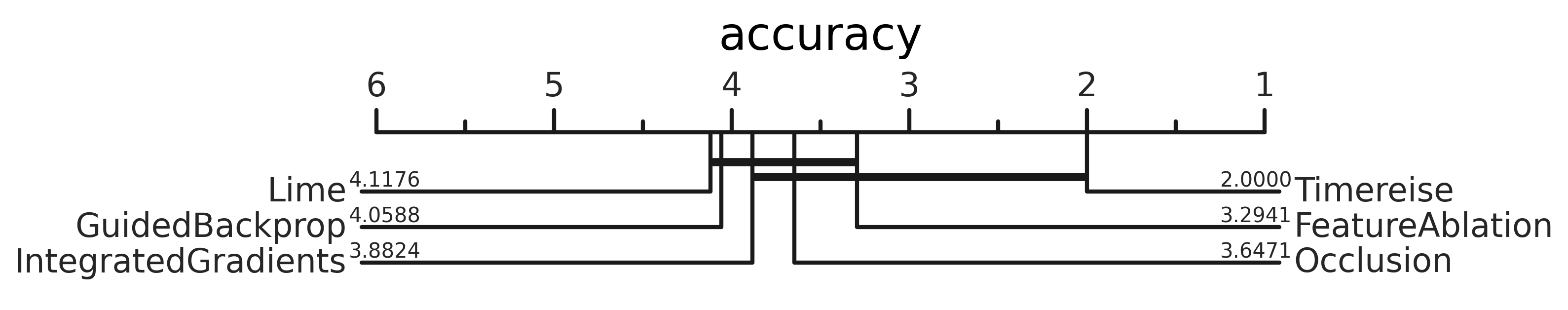}
\label{fig:del_ins_d}
}
\hfil
\subfloat[Insertion of important data points]{
\includegraphics[width=.96\linewidth]{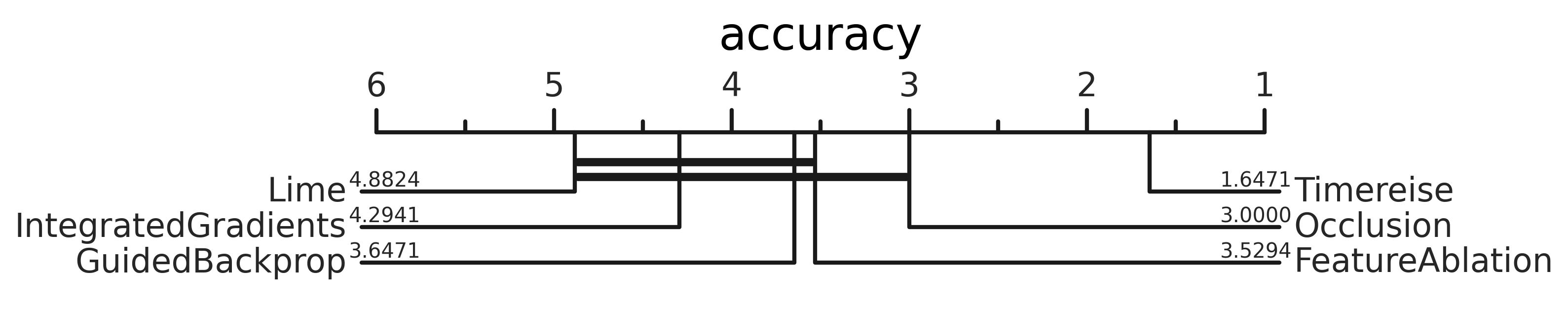}
\label{fig:del_ins_i}
}
\caption{\textbf{Deletion \& Insertion.} Critical difference diagram showing the average rank of each attribution method across all datasets. Ranking is based on the AUC using the accuracy. Perturbation-based approaches achieve better results.}
\label{fig:del_ins}
\end{figure}

\begin{table*}[!t]
\renewcommand{\arraystretch}{1.3}
\caption{\textbf{Deletion \& Insertion.} Sequential deletion of the most important points from the original input signal. Respectively, sequential insertion of the most important points starting with a samples consisting of mean values. Lower AUC scores are better for deletion. Higher AUC scores are better for insertion. AUC calculated using classification accuracy. TimeREISE outperforms any other methods concerning the deletion and achieves the best average for both deletion and insertion.} 
\label{tab:del_ins}
\centering
\scalebox{0.91}{
\begin{tabular}{l|r|r|r|r|r|r|r|r|r|r|r|r}
\multirow{2}{*}{\textbf{Dataset}} & 
\multicolumn{2}{c|}{\textbf{FeatureAblation~\cite{fisher2019all}}} & 
\multicolumn{2}{c|}{\textbf{GuidedBackprop~\cite{sundararajan2017axiomatic}}} & 
\multicolumn{2}{c|}{\textbf{IntegratedGrad.~\cite{springenberg2014striving}}} & 
\multicolumn{2}{c|}{\textbf{Lime~\cite{lime}}} & 
\multicolumn{2}{c|}{\textbf{Occlusion~\cite{zeiler2014visualizing}}} & 
\multicolumn{2}{c}{\textbf{Timereise (ours)}} \\
                         & del    & ins    & del    & ins    & del    & ins    & del    & ins    & del    & ins    & del    & ins    \\
\hline\hline
Anomaly                  & 0.7731 & 0.9737 & 0.7791 & 0.9597 & 0.7786 & 0.9624 & 0.7783 & 0.9473 & 0.7714 & 0.9739 & 0.7631 & 0.9867 \\
AsphaltPavementType      & 0.4073 & 0.8819 & 0.3930 & 0.8944 & 0.3940 & 0.8935 & 0.4622 & 0.8623 & 0.4171 & 0.8726 & 0.4135 & 0.8641 \\
AsphaltRegularity        & 0.5857 & 0.9954 & 0.5785 & 0.9960 & 0.5817 & 0.9964 & 0.6843 & 0.9871 & 0.5901 & 0.9929 & 0.5927 & 0.9833 \\
CharacterTrajectories    & 0.0856 & 0.8563 & 0.0807 & 0.8701 & 0.1091 & 0.8580 & 0.0785 & 0.8543 & 0.0878 & 0.8609 & 0.0401 & 0.8809 \\
Crop                     & 0.0998 & 0.3780 & 0.1402 & 0.3026 & 0.1404 & 0.2652 & 0.1096 & 0.3198 & 0.1583 & 0.3170 & 0.0628 & 0.5065 \\
ECG5000                  & 0.2104 & 0.8771 & 0.1876 & 0.8782 & 0.1208 & 0.8792 & 0.1294 & 0.8846 & 0.1176 & 0.8796 & 0.1015 & 0.9060 \\
ElectricDevices          & 0.3086 & 0.5393 & 0.3616 & 0.5718 & 0.3178 & 0.5244 & 0.3338 & 0.4971 & 0.3524 & 0.5914 & 0.2726 & 0.6957 \\
FaceDetection            & 0.5165 & 0.6760 & 0.2462 & 0.8065 & 0.5116 & 0.6660 & 0.6019 & 0.6308 & 0.5281 & 0.6691 & 0.0080 & 0.9968 \\
FordA                    & 0.4729 & 0.7816 & 0.4829 & 0.8207 & 0.4793 & 0.6834 & 0.4803 & 0.6731 & 0.4751 & 0.8493 & 0.3859 & 0.9436 \\
HandOutlines             & 0.3125 & 0.3630 & 0.3137 & 0.3289 & 0.3127 & 0.3432 & 0.3153 & 0.3201 & 0.3107 & 0.3911 & 0.3485 & 0.3607 \\
MedicalImages            & 0.1840 & 0.5884 & 0.1588 & 0.5645 & 0.1953 & 0.4518 & 0.1736 & 0.5622 & 0.1569 & 0.5883 & 0.1229 & 0.7125 \\
MelbournePedestrian      & 0.1579 & 0.5967 & 0.2071 & 0.5579 & 0.2733 & 0.4579 & 0.1767 & 0.6013 & 0.2363 & 0.4763 & 0.0979 & 0.6538 \\
NonInvasiveFetalECG      & 0.0424 & 0.1488 & 0.0454 & 0.0654 & 0.0405 & 0.0868 & 0.0462 & 0.0816 & 0.0422 & 0.2503 & 0.0894 & 0.4333 \\
PhalangesOutlinesCorrect & 0.4033 & 0.5072 & 0.4058 & 0.4347 & 0.4056 & 0.4437 & 0.4038 & 0.4288 & 0.4034 & 0.5616 & 0.2919 & 0.6171 \\
Strawberry               & 0.5827 & 0.7179 & 0.6141 & 0.7100 & 0.6428 & 0.7179 & 0.6397 & 0.7087 & 0.5958 & 0.7761 & 0.3882 & 0.7909 \\
UWaveGestureLibraryAll   & 0.1840 & 0.4243 & 0.1353 & 0.5260 & 0.1285 & 0.1452 & 0.1226 & 0.1782 & 0.1743 & 0.4669 & 0.0973 & 0.5379 \\
Wafer                    & 0.2740 & 0.7684 & 0.3441 & 0.8574 & 0.2603 & 0.8061 & 0.2324 & 0.8613 & 0.2642 & 0.7932 & 0.2002 & 0.8976 \\
\hline\hline
Average                  & 0.3295 & 0.6514 & 0.3220 & 0.6556 & 0.3348 & 0.5989 & 0.3393 & 0.6117 & 0.3342 & 0.6653 & \textbf{0.2516} & \textbf{0.7510} \\
\end{tabular}
}
\end{table*}

Figure~\ref{fig:del_ins} shows the critical difference diagrams of every attribution method. These were calculated using the AUC based on the achieved accuracy. In Figure~\ref{fig:del_ins_d} TimeREISE shows an outstanding performance compared to the other state-of-the-art methods with respect to the deletion of important data that affects the classifier performance. Another important finding is that the methods that utilize a window such as FeatureAblation and Occlusion show better performances concerning the deletion compared to methods that directly depend on the gradients such as GuidedBackprop and IntegratedGradients. However, Figure~\ref{fig:del_ins_i} highlights that the results are the opposite for the insertion task. One reason for is outcome is the smoothing applied to approaches that use a defined window. Gradient-based based method provide noisy and spiking attribution maps which are better suited for the insertion.

Table~\ref{tab:del_ins} shows the different results of the deletion and insertion for every individual dataset. Furthermore, the table provides the average scores achieved by the methods. TimeREISE shows a superior behavior in both the average deletion and insertion score. TimeREISE achieves the best (lowest) score for $13$ datasets and an average of $0.2516$. The second best approach concerning the average AUC score is GuidedBackprop with a score of $0.3220$ and two times the best performance. While TimeReise has the best average score for the insertion as well, it scores only two times the performance. GuidedBackprop achieves five times, IntegratedGradients four times and Lime three times the best score in the insertion task. However, the average score of TimeREISE is $0.7510$ compared to the second best of $0.6653$ for the Occlusion.

\subsection{Infidelity \& Sensitivity}
The Infidelity and Sensitivity proposed by Yeh et al.~\cite{yeh2019fidelity} cover significant and insignificant changes applied to the attribution and the input. The intuition behind the Infidelity is that a significant perturbation of the attribution map leads to a change in the prediction. Similarly, the Sensitivity is calculated using a insignificant change in the input sample. In the later case, it is mandatory to recompute the attribution map. For both, Infidelity and Sensitivity lower values are better. For the infidelity $1,000$ perturbations were computed for each of the $100$ samples and computed the averaged Infidelity value. In addition, $10$ perturbations for each of the samples were computed and their Sensitivity was calculated.

\begin{table}[!t]
\renewcommand{\arraystretch}{1.3}
\caption{\textbf{Infidelity.} Lower values correspond to better performance. The Method names are shortened by taking only the initial character. There are only insignificant differences between the methods.} 
\label{tab:infidelity}
\centering
\scalebox{0.76}{
\begin{tabular}{l|r|r|r|r|r|r}
\textbf{Dataset} & 
\textbf{F~\cite{fisher2019all}} & 
\textbf{G~\cite{sundararajan2017axiomatic}} & 
\textbf{I~\cite{springenberg2014striving}} & 
\textbf{L~\cite{lime}} & 
\textbf{O~\cite{zeiler2014visualizing}} & 
\textbf{T (ours)} \\
\hline\hline
Anomaly                  & 0.0233 & 0.0193 & 0.0158 & 0.0184 & 0.0222 & 0.0230 \\
AsphaltPavementType      & 0.2126 & 0.2126 & 0.2126 & 0.2127 & 0.2126 & 0.2124 \\
AsphaltRegularity        & 0.0045 & 0.0046 & 0.0046 & 0.0046 & 0.0045 & 0.0045 \\
CharacterTrajectories    & 0.1399 & 0.1397 & 0.1399 & 0.1399 & 0.1399 & 0.1396 \\
Crop                     & 0.2967 & 0.3081 & 0.3055 & 0.2966 & 0.3143 & 0.3032 \\
ECG5000                  & 0.0273 & 0.0272 & 0.0257 & 0.0210 & 0.0236 & 0.0242 \\
ElectricDevices          & 18.0869 & 18.1047 & 18.1130 & 18.1042 & 18.0854 & 18.1070 \\
FaceDetection            & 0.0002 & 0.0002 & 0.0002 & 0.0002 & 0.0002 & 0.0002 \\
FordA                    & 0.0118 & 0.0118 & 0.0116 & 0.0116 & 0.0118 & 0.0118 \\
HandOutlines             & 1.6914 & 1.7015 & 1.6932 & 1.6928 & 1.6938 & 1.6920 \\
MedicalImages            & 0.2492 & 0.2492 & 0.2472 & 0.2486 & 0.2490 & 0.2482 \\
MelbournePedestrian      & 1.2324 & 1.2833 & 1.3745 & 1.1959 & 1.3319 & 1.2301 \\
NonInvasiveFetalECG      & 51.7361 & 51.7288 & 51.7252 & 51.7228 & 51.7413 & 51.7072 \\
PhalangesOutlinesCorrect & 0.4394 & 0.4285 & 0.4360 & 0.4413 & 0.4403 & 0.4405 \\
Strawberry               & 0.4865 & 0.4783 & 0.4863 & 0.4849 & 0.4811 & 0.4851 \\
UWaveGestureLibraryAll   & 4.9995 & 4.9983 & 4.9922 & 4.9996 & 4.9992 & 4.9968 \\
Wafer                    & 0.0355 & 0.0356 & 0.0355 & 0.0356 & 0.0355 & 0.0352 \\
\hline\hline
Average                  & 4.6867 & 4.6901 & 4.6952 & 4.6842 & 4.6933 & 4.6859 \\
\end{tabular}
}
\end{table}

Starting with the Infidelity, the results shown in Table~\ref{tab:infidelity} emphasize that there is no significant difference between the different methods. Overall the average scores differ only by $0.011$ which is an insignificant difference. Across all dataets the methods perform similarly and it is not possible to create a critical difference diagram as the null hypothesis does hold. Interestingly, the Infidelity scores for the ElectricDevices and PhalangesOutlinesCorrect dataset are much lager compared to those of any other dataset. 

\begin{table}[!t]
\renewcommand{\arraystretch}{1.3}
\caption{\textbf{Sensitivity.} Lower values correspond to better performance. The Method names are shortened by taking only the initial character. Perturbation-based approaches show an superior performance.} 
\label{tab:sensitivity}
\centering
\scalebox{0.76}{
\begin{tabular}{l|r|r|r|r|r|r}
\textbf{Dataset} & 
\textbf{F~\cite{fisher2019all}} & 
\textbf{G~\cite{sundararajan2017axiomatic}} & 
\textbf{I~\cite{springenberg2014striving}} & 
\textbf{L~\cite{lime}} & 
\textbf{O~\cite{zeiler2014visualizing}} & 
\textbf{T (ours)} \\
\hline\hline
Anomaly                  & 0.0574 & 0.0747 & 0.1470 & 0.2591 & 0.0664 & 0.0522 \\
AsphaltPavementType      & 0.0292 & 0.2864 & 0.0358 & 0.4259 & 0.0274 & 0.0705 \\
AsphaltRegularity        & 0.0288 & 0.2797 & 0.0567 & 0.3664 & 0.0274 & 0.0028 \\
CharacterTrajectories    & 0.0199 & 0.0547 & 0.0705 & 0.1353 & 0.0174 & 0.0076 \\
Crop                     & 0.0808 & 0.1060 & 0.1702 & 0.1786 & 0.1307 & 0.0411 \\
ECG5000                  & 0.0301 & 0.0772 & 0.1218 & 0.1811 & 0.0248 & 0.0111 \\
ElectricDevices          & 0.2069 & 0.2608 & 0.6129 & 0.2622 & 0.1949 & 0.1696 \\
FaceDetection            & 0.0180 & 0.0204 & 0.0136 & 0.4722 & 0.0144 & 0.0048 \\
FordA                    & 0.0231 & 0.0384 & 0.0708 & 0.1690 & 0.0155 & 0.0147 \\
HandOutlines             & 0.0952 & 0.1545 & 0.1203 & 0.1249 & 0.0743 & 0.1175 \\
MedicalImages            & 0.0428 & 0.0680 & 0.1483 & 0.1754 & 0.0395 & 0.0406 \\
MelbournePedestrian      & 0.1667 & 0.1363 & 0.1684 & 0.2514 & 0.2176 & 0.0472 \\
NonInvasiveFetalECG      & 0.1142 & 0.1043 & 0.1543 & 0.1564 & 0.0869 & 0.1570 \\
PhalangesOutlinesCorrect & 0.0415 & 0.1442 & 0.1562 & 0.1212 & 0.0390 & 0.0574 \\
Strawberry               & 0.0486 & 0.0966 & 0.0506 & 0.1267 & 0.0515 & 0.0698 \\
UWaveGestureLibraryAll   & 0.0569 & 0.0535 & 0.2341 & 0.1778 & 0.0373 & 0.0381 \\
Wafer                    & 0.0252 & 0.0368 & 0.1299 & 0.1250 & 0.0141 & 0.0051 \\
\hline\hline
Average                  & 0.0638 & 0.1172 & 0.1448 & 0.2182 & 0.0635 & 0.0533 \\
\end{tabular}
}
\end{table}

The Sensitivity experiments are listed in Table~\ref{tab:sensitivity}. The results of these experiments show a significant difference between the methods. The best result was achieved by TimeREISE with a score of $0.0533$. The worst result was achieved by Lime with a score of $0.2182$ which is about four times larger then the score of TimeREISE. The overall finding was that the perturbation-based approaches are superior in case of Sensitivity compared to the gradient-based or others. This is the case as the gradient-based methods result in noisy attribution maps whereas the perturbation-based come up with smoothed maps based on a window of multiple features. This smoothing increases the robustness against small changes in the input.

\begin{figure}[!t]
\centering
\includegraphics[width=\linewidth]{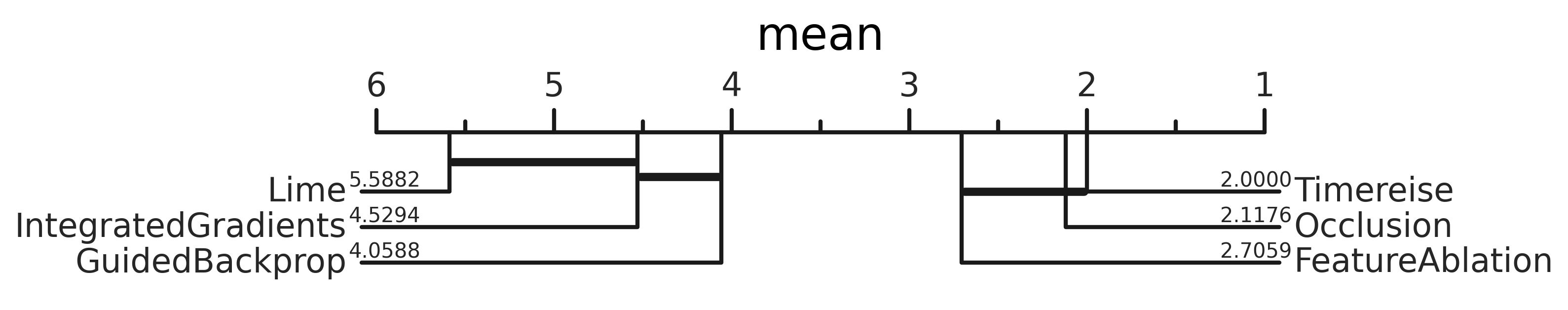}
\caption{\textbf{Sensitivity.} Critical difference diagram showing the average rank of each attribution method across all datasets. Ranking is based on the average Sensitivity. Perturbation-based approaches show an superior performance.}
\label{fig:sensitivity}
\end{figure}

In Figure~\ref{fig:sensitivity} the critical difference diagram across all datasets is provided. It shows the superior performance of the perturbation-based approaches compared to the other approaches. In addition, it highlights that TimeREISE is only slightly above the Occlusion method.

\subsection{Attribution Continuity}
Furthermore, this work calculated the Continuity proposed by Abdul et al.~\cite{abdul2020cogam}. The continuity is a measurement that bridges the correctness and the visual interpretability. The Continuity for each features was calculated as presented in Equation~\ref{eq:continuity} and took the mean for the overall evaluation between the methods. Lower values are better with respect to the cognitive load but might conflict with the exact correctness of the feature importance. 

\begin{equation}
    \label{eq:continuity}
    C = \sum_{c=0}^{C} \sum_{t=0}^{T-1} \mid S_{c,t} - S_{c,t+1} \mid
\end{equation}

\begin{table}[!t]
\renewcommand{\arraystretch}{1.3}
\caption{\textbf{Continuity.} Lower values correspond to better performance. The Method names are shortened by taking only the initial character. TimeREISE shows a superior performance.} 
\label{tab:continuity}
\centering
\scalebox{0.76}{
\begin{tabular}{l|r|r|r|r|r|r}
\textbf{Dataset} & 
\textbf{F~\cite{fisher2019all}} & 
\textbf{G~\cite{sundararajan2017axiomatic}} & 
\textbf{I~\cite{springenberg2014striving}} & 
\textbf{L~\cite{lime}} & 
\textbf{O~\cite{zeiler2014visualizing}} & 
\textbf{T (ours)} \\
\hline\hline
Anomaly                  & 0.1163 & 0.1444 & 0.1309 & 0.1390 & 0.0908 & 0.0473 \\
AsphaltPavementType      & 0.0792 & 0.0977 & 0.0770 & 0.0765 & 0.0450 & 0.0015 \\
AsphaltRegularity        & 0.0582 & 0.0703 & 0.0485 & 0.0525 & 0.0334 & 0.0008 \\
CharacterTrajectories    & 0.0264 & 0.0324 & 0.0368 & 0.0619 & 0.0243 & 0.0134 \\
Crop                     & 0.1282 & 0.1655 & 0.1952 & 0.1741 & 0.0985 & 0.0618 \\
ECG5000                  & 0.0682 & 0.1000 & 0.1004 & 0.0844 & 0.0505 & 0.0296 \\
ElectricDevices          & 0.2016 & 0.1840 & 0.1984 & 0.1950 & 0.0884 & 0.0350 \\
FaceDetection            & 0.0690 & 0.0745 & 0.0613 & 0.0331 & 0.0373 & 0.0161 \\
FordA                    & 0.0770 & 0.0819 & 0.0959 & 0.1530 & 0.0576 & 0.0083 \\
HandOutlines             & 0.0123 & 0.0183 & 0.0258 & 0.1501 & 0.0106 & 0.0015 \\
MedicalImages            & 0.0923 & 0.1043 & 0.1259 & 0.1076 & 0.0602 & 0.0371 \\
MelbournePedestrian      & 0.1804 & 0.1844 & 0.2217 & 0.1881 & 0.1264 & 0.1052 \\
NonInvasiveFetalECG      & 0.0224 & 0.0650 & 0.0753 & 0.1603 & 0.0197 & 0.0043 \\
PhalangesOutlinesCorrect & 0.1066 & 0.1187 & 0.1525 & 0.1416 & 0.0715 & 0.0496 \\
Strawberry               & 0.0720 & 0.0679 & 0.0785 & 0.1447 & 0.0676 & 0.0159 \\
UWaveGestureLibraryAll   & 0.0216 & 0.0557 & 0.0816 & 0.1629 & 0.0226 & 0.0038 \\
Wafer                    & 0.0924 & 0.0957 & 0.1418 & 0.1222 & 0.0557 & 0.0232 \\
\hline\hline
Average                  & 0.0838 & 0.0977 & 0.1087 & 0.1263 & 0.0565 & 0.0267 \\
\end{tabular}
}
\end{table}

\begin{figure}[!t]
\centering
\includegraphics[width=\linewidth]{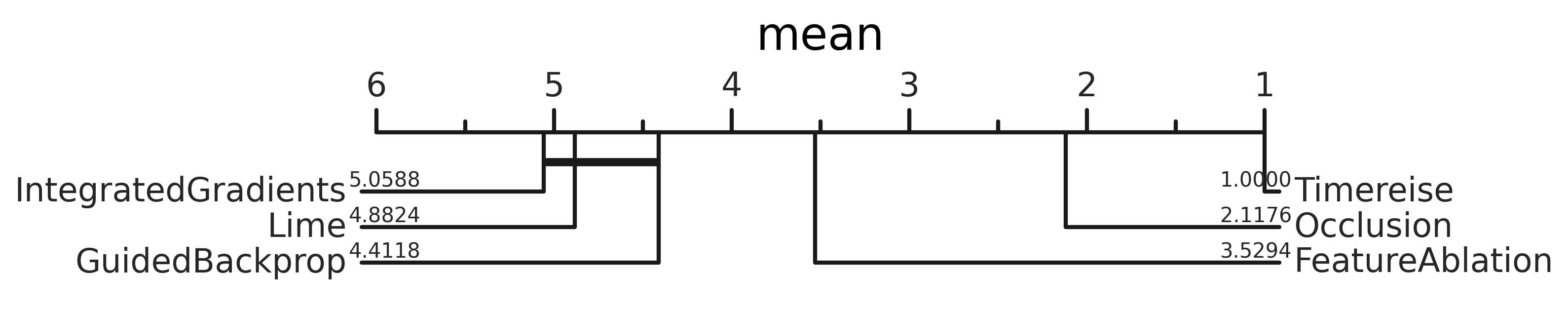}
\caption{\textbf{Continuity.} Critical difference diagram showing the average rank of each attribution method across all datasets. Ranking is based on the average Continuity. PErturbation-based approaches show a superior performance.}
\label{fig:continuity}
\end{figure}

\begin{figure*}[!t]
\centering
\subfloat[Anomaly detection dataset sample. Anomalies are defined as point anomalies represented by peaks.]{
\includegraphics[width=.9\linewidth]{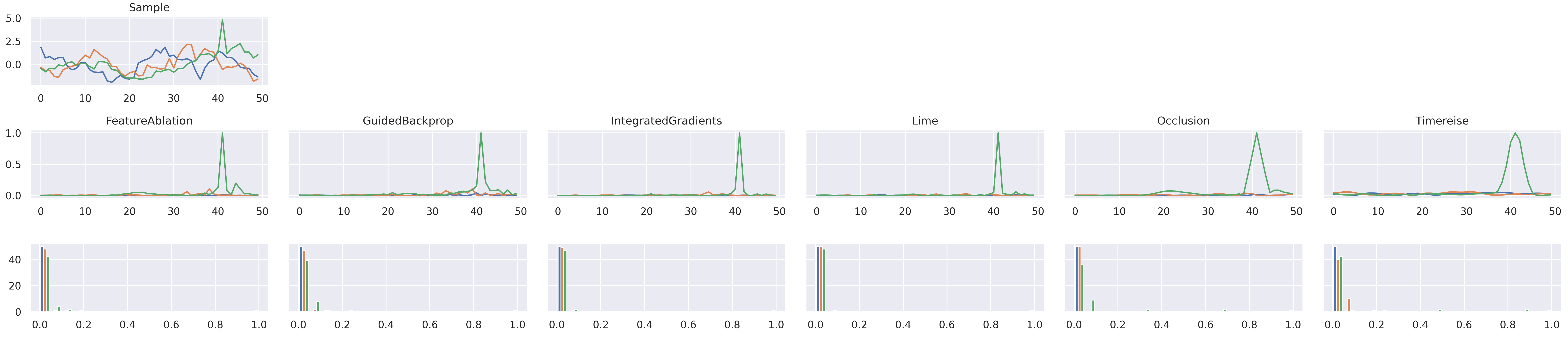}
\label{fig:attr_anomaly}
}
\hfil
\subfloat[ECG5000 data sample. Classification task that depends on a single channel.]{
\includegraphics[width=.9\linewidth]{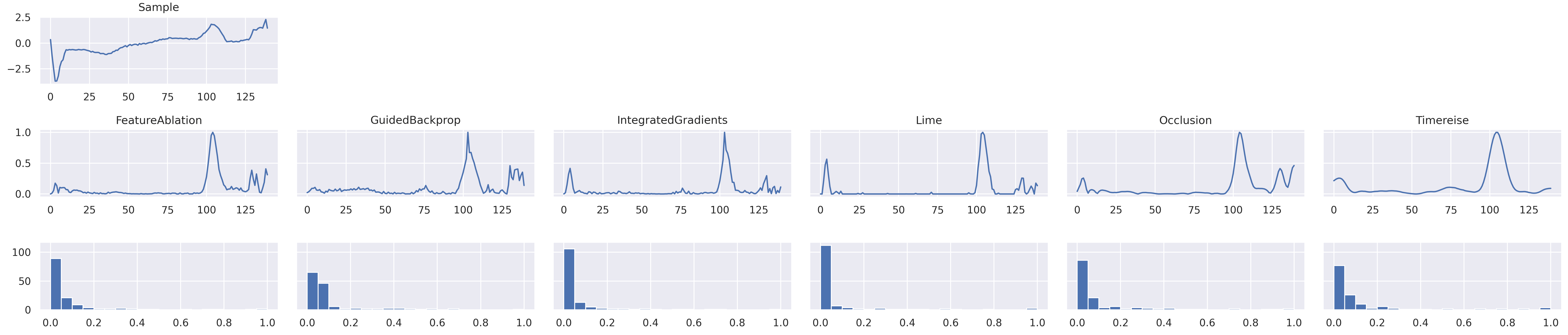}
\label{fig:attr_ecg}
}
\caption{\textbf{Attribution Maps.} Shows the attribution maps for a single sample. First row shows the original sample. Second row shows the actual attribution and third row shows the histogram of the attribution scores within the given map. Generally, a large amount of low values in the histogram relates to a good separation between relevnat and irrelevant features. Figure~\ref{fig:attr_anomaly} shows an anomalous sample in which the green peak corresponds to the anomalous signal part. Overall the attributions look similar except that TimeREISE is smoother compared to the other methods.}
\label{fig:attrs}
\end{figure*}

In Table~\ref{tab:continuity} we show the average Continuity of the attribution methods. Similar to the Sensitivity smaller values are better. Interestingly, the performance of the attribution methods is very similar to the Sensitivity. Again TimeREISE shows superior performance with a score of $0.0267$ compared to Occlusion as second best approach with a score of $0.0565$. The reason for the superior performance is the smooth mask design. The masks of TimeREISE are created on a downscaled sample and then they are upscaled using interpolation to the original input size. This results in smoother masks compared to Occlusion and FeatureAblation which utilize fixed windows.

Figure~\ref{fig:continuity} shows the corresponding critical difference diagram. It is intuitive that the Sensitivity defined as the change in prediction when the attribution method is applied to a slightly perturbed input and the Continuity the smoothness of the attribution method are connected to each other. However, it is interesting to observe the strong correlation between those two aspects.

\subsection{Visualization}
This section presents some interpretable attribution maps. The results highlight that TimeREISE produces smoother attribution maps while preserving the overall similar shape compared to the other attribution methods. TimeREISE builds a good compromise between the visual appearance that is strongly affected by the Continuity and the noise and the correctness of the feature importance values.

In Figure~\ref{fig:attrs} an attribution map of every evaluated attribution map is shown. The first Figure~\ref{fig:attr_anomaly} shows an anomalous sample of the Anomaly dataset. The anomaly is represented by the peak in the green signal in the first subplot. All methods successfully identify the peak as the most important part. However, the Occlusion and TimeREISE highlight that the neighborhood points of the peak are important. Whereas the intuition first suggests that only the peak should be highlighted this is not correct as changing the neighborhood points will influence the peak. Furthermore, it is visible that the attribution map provided by TimeREISE is very smooth compared to the other attributions while preserving the relevant information. 

In Figure~\ref{fig:attr_ecg} an attribution map for the ECG5000 dataset is shown. The results of all methods look similar to a certain degree. However, except TimeREISE the last part of the sequence is identified as features with some importance and the attribution maps include some noise. Specifically, the first negative peak in the signal is captured by the IntegratedGradients and Lime to be an important part. This is not the case for the remaining methods and changing this part or the last part has only minor effect on the prediction.

\begin{figure*}[!t]
\centering
\includegraphics[width=.9\linewidth]{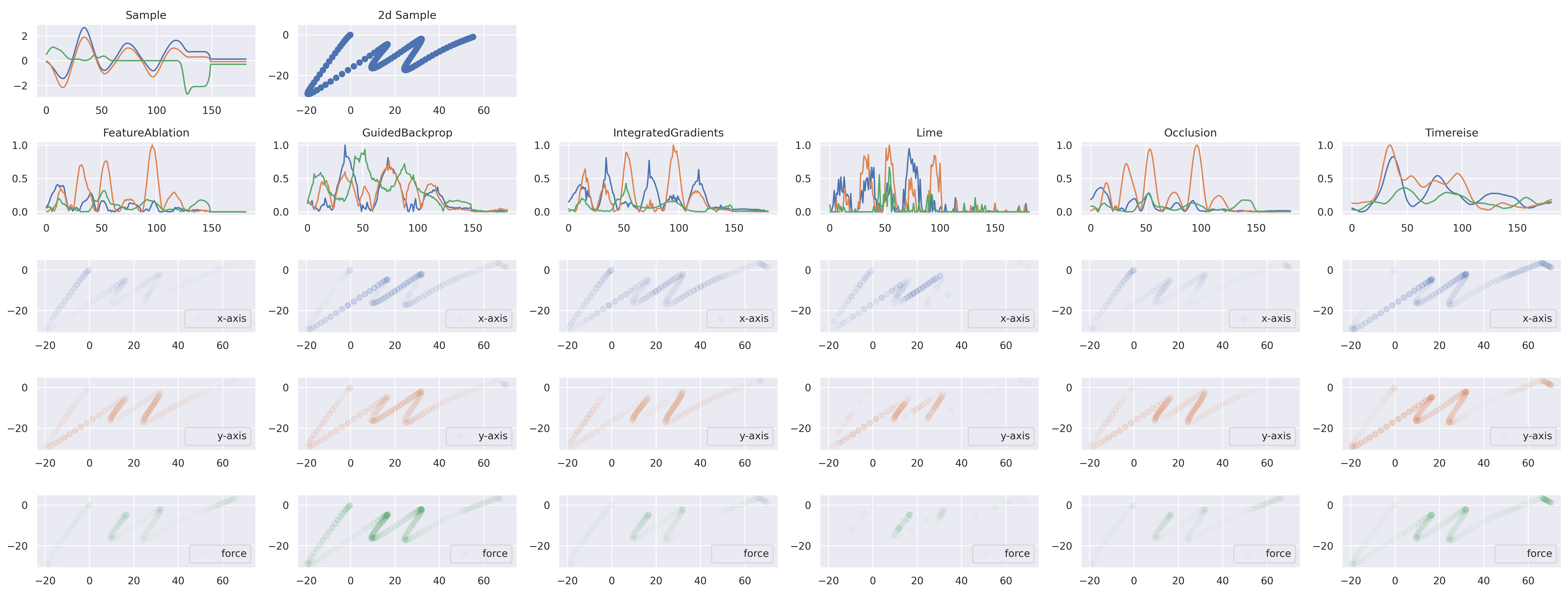}
\caption{\textbf{Explainable Attribution.} First row shows the time series of the character example 'm'. The right plot corresponds to the back transformation to the original 2d space. The second row shows the attribution results for each method. The subsequent rows show the importance applied to the character for the horizontal and vertical direction as well as for the force. TimeREISE provides a smooth attribution map and assigns importance to the force channel. Beside TimeREISE, only GuidedBackprop highlights the importance of the force channel.}
\label{fig:attr_char}
\end{figure*}

Figure~\ref{fig:attr_char} shows the results of the attribution applied to an interpretable character trajectory sample. The Figure presents the time series sample and its back transformation to 2d space. Furthermore, the attribution maps given in the second row show the smoothness of TimeREISE. One finding is that the horizontal and vertical movement are rated as more important by most methods and that the majority of important points occurs within the first $100$ timesteps. Interestingly, GuidedBackprop results in a surprisingly high relevance for the force. FeatureAblation and Occlusion show a low importance for both the vertical movement and the pressure.

\section{Conclusion}
\label{Conclusion}
This work shows that the novel attribution method TimeREISE is able to achieve excellent performance with respect to most of the evaluated metrics across all selected datasets. Precisely, the method outperforms other state-of-the-art attribution methods when it comes to the Continuity, Sensitivity, and causal metric. Specifically, the deletion scores when important data is occluded shows that the method provides superb performance. Furthermore, the paper has shown that the method provides smooth attribution maps that require significantly less effort to be interpreted. Considering the Infidelity, out method is on par with the state-of-the-art methods. Further, the theoretical runtime evaluation shows that the method has a significantly better scaling compared to methods that directly depend on the number of features and is applicable to any classifier. Another positive aspect is that the method does not depend on noisy gradients or internal classifier variables.

\iflong
\appendix
TODO
\else
\fi

\ifreview\else
\section*{Acknowledgment}
This work was supported by the BMBF projects SensAI (BMBF Grant 01IW20007) and the ExplAINN (BMBF Grant 01IS19074). We thank all members of the Deep Learning Competence Center at the DFKI for their comments and support.
\fi

\bibliographystyle{IEEEtran}
\bibliography{bibliography}

\end{document}